*Review*

# A Review of Mobile Mapping Systems: From Sensors to Applications

**Mostafa Elhashash** [1,3], **Hessah Albanwan** [1,2], and **Rongjun Qin** [1,2,3,4,*]

[1] Geospatial Data Analytics Lab, The Ohio State University, Columbus, USA
[2] Department of Civil, Environmental and Geodetic Engineering, The Ohio State University, Columbus, USA
[3] Department of Electrical and Computer Engineering, The Ohio State University, Columbus, USA
[4] Translational Data Analytics Institute, The Ohio State University, Columbus, USA
[*] Correspondence: qin.324@osu.edu

**Abstract:** The evolution of mobile mapping systems (MMSs) has gained more attention in the past few decades. MMSs have been widely used to provide valuable assets in different applications. This has been facilitated by the wide availability of low-cost sensors, the advances in computational resources, the maturity of the mapping algorithms, and the need for accurate and on-demand geographic information system (GIS) data and digital maps. Many MMSs combine hybrid sensors to provide a more informative, robust, and stable solution by complementing each other. In this paper, we present a comprehensive review of the modern MMSs by focusing on 1) the types of sensors and platforms, where we discuss their capabilities, limitations, and also provide a comprehensive overview of recent MMS technologies available in the market, 2) highlighting the general workflow to process any MMS data, 3) identifying the different use cases of mobile mapping technology by reviewing some of the common applications, and 4) presenting a discussion on the benefits, challenges, and share our views on the potential research directions.

## 1. Introduction

The need for regularly updated and accurate geospatial data has grown exponentially in the last decades. The geospatial data serves as an important source for various applications, including but not limited to indoor and outdoor 3D modeling, generation of geographic information system (GIS) data, disaster response high-definition (HD) maps, and autonomous vehicles. Such data collection has become possible by the continuous advances in mobile mapping systems (MMSs). MMS refers to an integrated system of mapping sensors mounted on a moving platform to provide the positioning of the platform while collecting geospatial data [1]. A typical MMS platform takes light detection and ranging (LiDAR) and/or high-resolution cameras as the primary sensors for acquiring data for objects/areas of interest, integrated with sensor suites for positioning and georeferencing, such as the global navigation satellite system (GNSS) and inertial measurement unit (IMU)). To perform accurate geo-referencing, traditional mobile mapping approaches require extensive post-processing, such as strip adjustment of point clouds scans, or bundle adjustment (BA) of images using ground control points (GCPs), during which manual operations may be required to clean noisy data and unsynchronized observations. Recent trends of MMSs aim to perform direct georeferencing that leverages the capabilities of a multi-sensor platform [2,3], to minimize the human interventions during the data collection and processing. The automation has been further strengthened to use machine learning/artificial intelligence to perform online/offline object extraction and mappings, such as traffic lights and road sign extraction [4-6].

Mobile mapping technology has experienced significant developments in the past few decades with algorithmic advances in photogrammetry, computer vision, and robotics [7]. In addition, the increased processing power and storage capacity further facilitate the collection speed and data volumes [8]. The applications and systems are further strengthened by the availability and a diverse set of low-cost survey sensors with various specifications, making mobile mapping more flexible to acquire data in complex environments (e.g., tunnels, caves, and enclosed spaces), with lower cost and fewer labors [9]. Typically, commercial MMSs can be classified based on their hosting platforms into handheld, backpack, trolley, and vehicle-based. Some platforms are designed to work indoors only without relying on GNSS, while others can work indoors and outdoors. Mobile mapping technology has gained more attention when it was adopted by companies such as Google and Apple [10,11] to be used for various applications, including navigation, virtual and augmented reality [12].

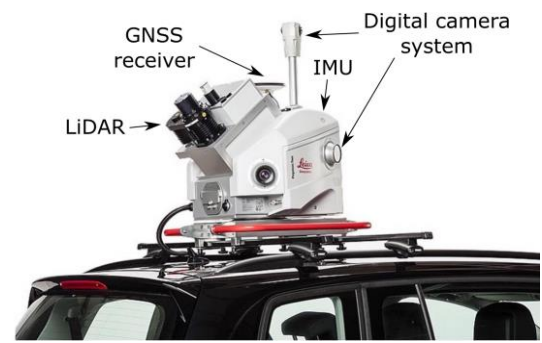

**Figure 1.** An example of an MMS: a vehicle-mounted mobile mapping platform consisting of different positioning and data collection sensors to generate an accurate georeferenced 3D map of the environment. We show the main sensors of the Leica Pegasus: Two Ultimate as an example. Photo courtesy of Leica Geosystems [13].

An example of a vehicle-based MMS (Leica Pegasus: Two Ultimate [13]) and its typical sensor suites are shown in Figure 1. It consists of both data acquisition sensors and positioning sensors. The data acquisition sensors primarily consist of a calibrated LiDAR and digital camera suite, where the LiDAR sensor is capable of producing 1,000,000 points/second and the camera suite captures 360 horizontal field of view (FoV) to provide both the texture/color information and the stereo measurement should this be needed. The positioning sensors include a GNSS receiver that provides the global positional information, with an additional IMU and distance measuring instrument (DMI) that obtain odometry information for integrated position correction. These positioning systems are required to be calibrated in their relative positions and play a vital role to generate globally consistent point clouds.

Despite the existence of a few mobile mapping technologies in the market, the technology landscape of MMS is highly disparate, as there is no single and standard MMS that is widely used in the mapping community. Most of the existing MMSs are customized using different sensor suites at different grades of integration, thus each has its pros and cons. Previous studies mostly focused on comparative studies among certain devices [1,14-19] or targeted systems for specific application scenarios (e.g., indoors or outdoors) [20-22]. Due to the rapid development of imaging, LiDAR, positioning sensors, and onboard computers, the updated capability of these essential components of an MMS, may not be fully reflected through a few integrated systems, thus making these studies less informative. There is generally a lack of study that covers the comprehensive landscape of sensor suites and the respective MMSs. In this paper, we focus on providing a meta-review of sensors and platforms tasked for ground-level 3D mappings, as well as the techniques needed to integrate these sensors as suites for MMS in different application scenarios. This review intends to provide an update on sensors, MMSs with different hosting platforms, and extended applications of MMSs, to not only serve researchers in the field of mobile mapping with updated background information but also for practitioners on critical factors of concern when planning to customize an MMS for specific applications. We will highlight the main steps from data acquisition to refinement, and discuss some of the most common challenges and considerations of MMS.

## 2. Paper Scope and Organization

This paper provides a comprehensive review of the MMS technology, where we provide a thorough discussion covering mobile mapping from the sensors, and software to their applications. We thoroughly discuss the types of the different sensors, their practical capabilities, and limitations, as well as methods of fusing these sensory data. We then describe the main platforms that are used in the mapping tasks of different application scenarios (e.g., indoor and outdoor applications). In addition, we discuss the main stages of processing MMS data including preprocessing, calibration, and refinement. In order to assess the benefits of an MMS in practice, we examine a few of the most important applications that widely use mobile mapping technology. Finally, for the benefit of future works and directions, we highlight the main considerations and challenges in an open discussion.

The rest of this paper is organized as follows: Section 3 provides a detailed review of essential positioning and data collection sensors in MMSs; Section 4 presents the different MMS hosting platforms based on their application scenarios (i.e., vehicle-mounted systems, handheld, wearable, and trolley-based); Section 5 presents the workflow to process MMS data from acquisition to algorithms for fusing their observations and refinement; Section 6 introduces the enabled applications using MMS on mapping and beyond. Finally, Section 7 concludes this review and discusses the future trends.

## 3. An Overview of Sensors in Mobile Mapping Systems

The positioning and data collection sensors are two classes of essential components of a typical MMS, as depicted in Figure 1. The positioning sensors are used to obtain the geographical positions and motion of the sensors, which are used to geo-reference the collected 3D data. Examples of these sensors include GNSS, IMU, DMI (i.e., odometers), etc. To achieve statistically more accurate positioning, the measurements from these sensors are usually jointly used through fusion. Addition fusion can be also performed between the position and the navigation cameras. The sensor fusion solution for positioning is nowadays standard, as neither the GNSS receiver nor the IMU/DMI alone, can provide sufficiently accurate and reliable measurements for navigating mobile platforms. GNSS measurements are usually subject to the signal strength variation in different environments, for example, we can obtain a strong signal in open spaces and weak or no signals in tunnels or indoors, which can lead to a loss of information. On the other hand, the IMU and DMI are subject to a significant accumulation of errors and thus are often used as supplemental observations for navigation when GPS data are available.

The data collection sensors mostly consist of LiDAR and digital cameras, providing raw 3D/2D measurements of the surrounding environments. The 3D measurements of an MMS mostly rely on LiDAR sensors, while the images are mostly used to provide the colorimetric/spectral information [20]. With the development of advanced dense image matching methods [23-27], these images are also collected stereoscopically to provide additional dense measurements for 3D data fusion. In the following subsections, we provide an overview of the positioning and data collection sensors, as well as the respective sensor fusion approaches.

*3.1. Positioning Sensors*

As mentioned above, typical positioning sensors include GNSS receiver, IMU, and DMI. Their patterns of errors are complementary to each other, thus in modern MMS, they are mostly used through a sensor fusion solution to provide accurate positioning information up to centimeter-level [28]. Nevertheless, their individual measurement accuracies are still decisive to the accuracy of the resulting 3D maps. An overview of the positioning sensors is shown in Table 1. In the following subsections, we will discuss the three main positioning sensors GNSS, IMU, and DMI.

**Table 1.** Positioning sensors overview.

| Sensor | Description | Benefits | Limitations |
|---|---|---|---|
| GNSS receiver | The signals from orbiting satellites are utilized by the GNSS receiver to compute the position, velocity, and elevation. Some examples include GPS, GLONASS, Galileo, and BeiDou. | • No/less accumulation of errors due to its dependence on external signals.<br>• Data collected under a global reference coordinate system (e.g., WGS84). | • Signal inaccessible in complex urban regions e.g., tall buildings, trees, tunnels, indoor environments, etc.<br>• Requires post-processing using DGPS and RTK-GPS to minimize errors from receiver's noise, pseudo-range, carrier phase, doppler shifts, atmospheric delays, etc. |
| IMU | IMU is an egocentric sensor that records the relative position of the orientation and directional acceleration of the host platform. | • Capable of navigating in all environments, such as indoors, outdoors, tunnels, caves, etc.<br>• A necessary supplemental data source for urban environments where GPS is unstable. | • Requires consistent calibration and a reference to avoid drift from the true position.<br>• Limited to short-range navigation. |
| DMI | A supplementary positioning sensor measures the traveled distance of the platform, i.e., information derived from a speedometer | • A supplemental sensor to provide additional data points to alleviate accumulation errors of IMU sensors. | • Requires calibration and provides only distance information (1 degree of freedom). |

### 3.1.1. Global Navigation Satellite System Receiver

The GNSS receiver is a primary source to estimate the absolute position, velocity, and elevation in open areas referenced to a global coordinate system (e.g., WGS84). It passively receives signals from a minimum of four different navigational satellite systems and performs trilateration to calculate its real-time positions. Since it depends on an external source of signal, the GNSS often exhibits less or no accumulation errors. These satellite systems mainly refer to the GPS developed by the United States, the GLONASS (Globalnaya Navigatsionnaya Sputnikovaya Sistema) developed by Russia, Galileo built by the European Union, and BeiDou developed by China [29]. The raw observations (pseudo-range, carrier phase, doppler shifts, etc.) from the chipset of the receiver with its solver often give a positional error at the meter level, depending on the chipsets and antenna (e.g., single/dual frequencies) [30]. High-tier MMSs often use augmented GPS solutions, such as Differential GPS (DGPS) or Real-Time Kinematic GPS (RTK-GPS), to improve the positioning accuracy to decimeters and centimeters levels (can achieve an accuracy of 1 cm [31]). DGPS uses a code-based measure and can operate with single-frequency receivers without initialization time, while RTK-GPS uses carrier-phase measures, thus it requires dual-frequency receivers and takes about one minute to initialize (for fixing the wavenumbers) [19]. Both DGPS and RTK-GPS rely on a network of reference stations to a surveyed point at its vicinity, to apply corrections to eliminate various errors such as ionosphere delays and other unmodeled errors. The traditional DGPS method achieves sub-meter accuracy in the horizontal position, while with much more advanced techniques and solvers, the RTK-GPS, as a type of DGPS, can achieve centimeter-level accuracy in three dimensions. However, these achievable accuracy measures are conditioned to open areas, while when collecting 3D data in dense urban areas with tall buildings or indoor environments, the GNSS signal can be largely impacted by the occlusions and the resulting measurements can be inaccurate [32], thus, it requires other complimentary sensors when operating under such conditions. In general, the positioning platform of an MMS is expected to achieve an accuracy of 5-50 mm at speed that can reach the maximum speed of highways (120-130 km/h) when considering the integration of complementary sensors.

### 3.1.2. Inertial Measurement Unit

IMU is an egocentric sensor that records the relative position of the orientation and directional acceleration of the host platform. Its positional information can be calculated through dead reckoning approaches [33,34]. Unlike GNSS, it does not require links to external signal sources, and it records relative positions with respect to a reference to the starting point (usually can be dynamically provided by GNSS in open fields). Like many other egocentric navigation methods, it suffers from accumulation errors leading to oftentimes significant drifts to its true positions. To be more specific, an IMU consists of an accelerometer and gyroscope to sense the acceleration and the angular velocity. These raw measurements are fed into an onboard computing unit to apply the dead reckoning algorithm to provide real-time positioning, thus the IMU and the computing unit with the algorithm as a whole is also called an inertial navigation system (INS). The grade/quality of IMU sensors can be differentiated by the type of gyroscopes: the majority of the light-weight and consumer-grade IMUs use Microelectromechanical Systems (MEMS) that have a low-cost but poor precision with large drift error (often 10-100°/h [35]) [3,36], and higher grade systems for precise navigation use a more sized but accurate gyroscopes such as ring laser and fiber optic gyroscopes, which can reach to a drift error less than 1°/h [35]. IMU can work in GPS-denied environments indoors, outdoors, and in tunnels, while given its nature of dead reckoning navigation, the measurements will only be accurate for a relatively short period in reference to the starting point. Since GNSS provides reasonable accuracy in an open area and measures do not have error accumulations as the platform moves, it often integrates IMU as additional observations, which, as a standard approach, provides more accurate positional information in complex environments mixed with both open and occluded surroundings [37].

### 3.1.3. Distance Measuring Instrument

The DMI generally refers to instruments that measure the traveled distance of the platform. In many cases, DMI is alternatively referred to as the odometer or wheel sensors for MMS based on vehicles or bikes. It computes the distance based on the number of cycles the wheel rotates. Since DMI only measures distance, it is often used as supplementary information to GNSS/IMU as an effective means to reduce the accumulated errors and constrain the drift from IMU in GPS-denied environments such as tunnels [38]. It requires calibration before use, which measures distance, velocity, and acceleration.

### 3.2. Sensors for Data Collection

Data collection sensors are another major component in the MMS to collect 3D data. They typically refer to sensors such as LiDAR and high-resolution cameras that provide both geometry and texture information. They require constant georeferencing using the position and orientation information provided by the positioning sensors to link the 3D data to the world coordinate system. In this section, we introduce the LiDAR and imaging systems (i.e., cameras), where we describe their functions, types, benefits, challenges, and limitations, and provide an example of a system that represents the status quo.

### 3.2.1. Light Detection and Ranging (LiDAR)

LiDAR, known as Light Detection and Ranging, is an optical instrument that uses directional laser beams to measure the distances and locations of objects. It provides individual and accurate point measurement on a 3D object, thus many of these measurements together constitute information about the shape and surface characteristics of objects in the scene. It has many wanted features in the 3D model, as it is highly accurate, can acquire dense 3D information in a short time, invariance to illumination, and can partially penetrate sparse objects like a canopy. LiDAR itself is still an instrument measuring relative locations, thus it requires a suite of a highly accurate and well-calibrated navigation system to retrieve global 3D points, the installation and cost of which, in addition to the already expensive LiDAR sensor, make it still a high-cost collection means.

The concept of using light beams for distance measurements has been used since 1930 [39]. Since the invention of the laser in 1960, LiDAR technology has experienced rapid development [40] and has been very popular in tasks like accurate mapping and autonomous driving applications [41]. Nowadays, there are many commercially available LiDAR sensors either for surveying or automotive applications. Typically, survey-grade LiDAR achieves a range accuracy at millimeter-level (usually 10-80 mm), such as RIEGL VQ-250, VQ-450 [42], and Trimble MX9 and MX50 [43]. Relatively lower-grade LiDAR sensors (with lower cost) achieve a range accuracy at a centimeter-level (usually 1-8 cm), generally satisfying the applications for obstacle avoidance and object detection, are often used in autonomous driving platforms given its good tradeoff between cost and performance. Examples of such LiDAR sensors include Velodyne [44], Ouster [45], Luminar Technology [46], and Innoviz Technologies [47]. In fact, the level of accuracy of different grades of LiDAR sensors and the cost are the main decisive factors when considering the choice of a LiDAR sensor, since there is a large gap in the cost of both grades, with survey-grade LiDARs often cost hundreds of thousands of US dollars (at the time of this publication), while the relatively lower-grade ones are about ten times cheaper.

LiDAR sensors can be categorized based on their collecting principle, into three main categories: rotating, solid-state, and flash. Rotating LiDAR uses a rotating mirror spinning for 360 degrees re-directing laser beams, and usually has multiple beams where each beam illuminates one point at a time. The rotating LiDAR is the most commonly used in MMS, since based on its nature of "rotating", it provides large FoV, high signal-to-noise ratio, and dense point clouds [48]. Solid-state LiDAR usually uses MEMS mirrors embedded in a chip [49] where the mirror can be controlled to follow a specific trajectory or optical phased arrays to steer beams [50], thus it is "solid" and does not possess any moving parts in the sensor. The Flash LiDAR [51] usually illuminates the entire FoV with a wide beam in a single pulse. Analogous to a camera with a flash, a flash LiDAR uses a 2D array of photodiodes to capture the laser returns, which are finally processed to form 3D point clouds [48,52]. Typically, a Flash LiDAR often has a limited range (less than 100 m) as well as a limited FoV, constrained by the sensor size. Although LiDAR is primarily used to generate point clouds, it can also be used for localization purposes through different techniques such as scan matching [53-55]. The extractable information can be further enhanced by deep neural networks for semantic segmentation and localization [56,57]. However, like many other methods, LiDAR sensors can provide relatively accurate range measurements, their performance deteriorates significantly in hazardous weather conditions such as heavy rain, snow, and fog.

Table 2 shows an example of several existing LiDAR sensors along with their technical specifications in terms of the range, accuracy, number of beams, FoV, resolution, point per second, and refresh rate. Generally, the choice of sensors depends on the application and the characteristics of the moving platform (e.g., the speed, payload, etc.). As mentioned above, most MMSs rely on rotating LiDAR sensors, but they often come at a high cost as compared to other categories. Therefore, using solid-state LiDAR in an MMS is now a promising direction since its cost is lower than the rotating-based LiDAR. When the vehicle speed is high, there is less time to acquire data and hence more beams are needed to ensure that the object of interest is measured by sufficient points [44,58]. For instance, a 32 beams LiDAR could be sufficient for a vehicle moving at a speed of 50-60 km/h, and a LiDAR with 128 beams is recommended for higher speeds up to 100-110 km/h so that the acquired data will have an adequate resolution. The operating range of a LiDAR can be also important and should be considered on an application basis (e.g., long-range LiDAR may be unnecessary for indoor applications.). In general, the cost usually increases roughly by a factor of 1.5-2 when the number of beams is doubled, which is also positively correlated to the operating range.

**Table 2.** Specifications of different LiDAR sensors.

| | Company | Model | Range (m) | Range accuracy (cm) | Number of beams | Horizontal FoV (°) | Vertical FoV (°) | Horizontal resolution (°) | Vertical resolution (°) | Points per second | Refresh rate (Hz) |
|---|---|---|---|---|---|---|---|---|---|---|---|
| **Rotating** | RIGEL | VQ-250 | 1.5-500 | 0.1 | — | 360 | — | — | — | 300,000 | — |
| | | VQ-450 | 1.5-800 | 0.8 | — | 360 | — | — | — | 550,000 | — |
| | Trimble | MX50 laser scanner | 0.6-80 | 0.2 | — | 360 | — | — | — | 960,000 | — |
| | | MX9 laser scanner | 1.2-420 | 0.5 | — | 360 | — | — | — | 1,000,000 | — |
| | Velodyne | HDL-64E | 120 | ±2 | 64 | 360 | 26.9 | 0.08 to 0.35 | 0.4 | 1,300,000 | 5 to 20 |
| | | HDL-32E | 100 | ±2 | 32 | 360 | 41.33 | 0.08 to 0.33 | 1.33 | 695,000 | 5 to 20 |
| | | Puck | 100 | ±3 | 16 | 360 | 30 | 0.1 to 0.4 | 2.0 | 300,000 | 5 to 20 |
| | | Puck LITE | 100 | ±3 | 16 | 360 | 30 | 0.1 to 0.4 | 2.0 | 300,000 | 5 to 20 |
| | | Puck Hi-Res | 100 | ±3 | 16 | 360 | 20 | 0.1 to 0.4 | 1.33 | 300,000 | 5 to 20 |
| | | Puck 32MR | 120 | ±3 | 32 | 360 | 40 | 0.1 to 0.4 | 0.33 (min) | 600,000 | 5 to 20 |
| | | Ultra Puck | 200 | ±3 | 32 | 360 | 40 | 0.1 to 0.4 | 0.33 (min) | 600,000 | 5 to 20 |
| | | Alpha Prime | 245 | ±3 | 128 | 360 | 40 | 0.1 to 0.4 | 0.11 (min) | 2,400,000 | 5 to 20 |
| | Ouster | OS2-32 | 1 to 240 | ±2.5 to ±8 | 32 | 360 | 22.5 | 0.18 | 0.7 | 655,000 | 10, 20 |
| | | OS2-64 | 1 to 240 | ±2.5 to ±8 | 64 | 360 | 22.5 | 0.18 | 0.36 | 1,311,000 | 10, 20 |
| | | OS2-128 | 1 to 240 | ±2.5 to ±8 | 128 | 360 | 22.5 | 0.18 | 0.18 | 2,621,000 | 10-20 |
| | Hesai | PandarQT | 0.1 to 60 | ±3 | 64 | 360 | 104.2 | 0.6° | 1.45 | 384,000 | 10 |
| | | PandarXT | 0.05 to 120 | ±1 | 32 | 360 | 31 | 0.09, 0.18, 0.36 | 1 | 640,000 | 5, 10, 20 |
| | | Oandar40M | 0.3 to 120 | ±5 to ±2 | 40 | 360 | 40 | 0.2, 0.4 | 1, 2, 3, 4, 5, 6 | 720,000 | 10, 20 |
| | | Oandar64 | 0.3 to 200 | ±5 to ±2 | 64 | 360 | 40 | 0.2, 0.4 | 1, 2, 3, 4, 5, 6 | 1,152,000 | 10, 20 |
| | | Pan-dar128E3X | 0.3 to 200 | ±8 to ±2 | 128 | 360 | 40 | 0.1, 0.2, 0.4 | 0.125, 0.5, 1 | 3,456,000 | 10, 20 |
| **Solid-state** | Luminar | IRIS | Up to 600 | — | 640 lines/second | 120 | 0-26 | 0.05 | 0.05 | 300 points/square degree | 1 to 30 |
| | Innoviz | InnovizOne | 250 | — | — | 115 | 25 | 0.1 | 0.1 | — | 5 to 20 |
| | | InnovizTwo | 300 | — | 8000 lines/second | 125 | 40 | 0.07 | 0.05 | — | 10 to 20 |
| **Flash** | LeddarTech | Pixell | Up to 56 | ±3 | — | 117.5 ± 2.5 | 16.0 ± 0.5 | — | — | — | 20 |
| | Continental | HFL110 | 50 | — | — | 120 | 30 | — | — | — | 25 |

"—" indicates that the specifications were not mentioned in the product datasheet

### 3.2.2. Imaging Systems and Cameras

Imaging systems like cameras are one of the most popular types of sensors for data collection due to their low-cost and ability to provide high-resolution texture information. Cameras are usually mounted on the top or front of the moving platform to capture information about the surrounding environment. They are intended to acquire many images at a high frame rate i.e., 30-60 frames per second. Cameras are useful to serve a few main purposes. First, recovering the geometry of the scene, which is usually obtained through stereoscopic/binocular cameras that process a pair of overlapping images to recover the depth information using stereo-dense image matching approaches [24-26]. Second, obtaining the textures of objects in the scene; a camera records photons of the object at different spectral frequencies that provide rich and critical information about the object's natural appearance, thus can be used to build panoramic and geotagged images, photo-realistic models. Third, the texture information encodes critical semantics of the object, thus can be used to detect static objects such as traffic lights, stop signs, marks, road lanes, and detect moving objects such as pedestrians and cars, which is becoming gradually applicable as the modern deep learning methods developed to tackle such problems [4,59,60].

There are many types of camera sensors and configurations used in MMSs, depending on their intended use as described before. Examples include monocular cameras, binocular cameras, RGB-D cameras, multi-camera systems (e.g., ladybug), fisheye, etc. A summary of different camera types is shown in Table 3. The monocular cameras (low-cost cameras) provide a series of single RGB images without any additional depth information, which is often used to collect high-resolution and geotagged images or panoramas [11]. However, they cannot be used to recover 3D scale or generate high accurate 3D points. The binocular cameras, on the other hand, consist of two cameras capturing synchronized stereo images to recover depth and scale with an additional computational cost through stereo-dense image matching techniques [24-26]. The performance and accuracy of the 3D information depend on the selection of the stereo-dense image matching method. In many cases, mapping solutions may rely on RGB-D cameras (e.g., Kinect [61], Intel RealSense D435 [62]) which can provide both RGB images and depth images (through structured light) simultaneously, they are mostly used in indoor settings due to their limited range. Integrating LiDAR with RGB-D images can yield highly accurate 3D information; however, this may require pre-compensation for the uncertainties in the RGB-D images from random noise or occlusions. Due to the compact/cluttered environment, an MMS often includes a wide FoV or even 360° panoramic camera, which is usually achieved via the multi-camera system that uses a group of synchronized cameras sharing the same optical center (e.g., FLIR Ladybug5+ [63]). Panoramic images additionally facilitate the integration with LiDAR scanners, providing 100% overlap between the image and LiDAR point clouds (e.g., those from rotating LiDAR). As a result, the panoramic images are suitable for street mapping applications. As a lower-cost alternative, a fisheye camera aims to provide an image with extended FoV from a single camera, it has a spherical lens that can provide more than 180° FOV, this is although cheaper, may come with the cost of image distortions in scale, geometry, shape, and illumination, requiring additional and slightly more complex calibrations.

As the modern MMS benefits from various cameras in providing additional information, it comes with a few added complexities. First, it captures images using the reflected light off of objects, which makes it sensitive to the illumination of the environment such as the high dynamic range of the scene (between sky and ground), and hazy weather conditions [64,65]. Second, cameras, and multi-camera systems, require calibration to reduce different types of distortions [66]. Third, moving platforms require high framerate cameras to leverage the speed, image quality, and resolution [67].

**Table 3.** Camera sensor overview

| Type | Description | Benefits | Limitations |
|---|---|---|---|
| Monocular | Single-lens camera | • Low cost.<br>• Provides a series of single RGB images to collect high-resolution and geotagged images or panoramas. | • Cannot recover 3D scale without additional sensors.<br>• Camera networks suboptimal to generate highly accurate 3D points |
| Binocular | Two collocated cameras with known relative orientation capturing overlapping and synchronized image | • Can provide depth and scale of objects the scene.<br>• Provides better accuracy integrated with LiDAR sensor. | • Performance and accuracy may depend on the algorithm used to compute the 3D information.<br>• |
| RGB-D | Cameras that capture RGB and depth images at the same time | • Simultaneous data acquisition.<br>• Provides high accuracy when integrated with LiDAR. | • Depth image sensitive to occlusions.<br>• Low range.<br>• The depth image may include some uncertainties and errors. |
| Multi-camera system | A spherical camera system with multiple cameras that can provide a 360° field of view | • Panoramic view showing the entire scene.<br>• Suitable for street mapping applications. | • Require large storage to save images in real-time.<br>• Must be properly calibrated to assure alignment of images and minimum distortions. |
| Fisheye | Spherical lens camera that has more than 180° field of view | • Provides wide coverage of the scene allowing capturing the scene with fewer images. | • Lens distortions.<br>• Non-projective transformation.<br>• Requires rigorous calibration. |

## 4. Mobile Mapping Systems and Platforms

There are a few factors that determine the type of sensor and platform to be used for MMS tasks. These factors include available sensors, project budget, technical solutions, processing strategies, and scene contents (i.e., indoor or outdoor). These help to determine the type of available sensors (e.g., with/without GPS) and the accessible platforms (e.g. vehicle-mounted or backpack, etc.). For example, in indoor environments there is no access to GPS signals or vehicles, thus, must adopt alternative solutions.

In general, we consider broadly categorizing the MMS platforms into traditional vehicles and non-traditional lightweight/portable mapping devices. Traditional vehicle-based MMS primarily operates on main roads collecting city or block-level 3D data. The non-traditional portable devices, such as backpack/wearable, handheld systems, or trolley-based depending on their application and task, can be used both outdoors or indoors in GPS-denied environments. For outdoor applications, these means of mapping is primarily used to complement vehicle-based system, by mapping narrow streets and areas that cannot be accessed by vehicles [68,69]. For indoor or GPS-denied environments, the sensor suites may be significantly different from those used outdoors, for example, they may primarily rely on INS or visual odometry for positioning [70,71]. To be more specific, in this section, we introduce four typical MMS platforms that offer mapping solutions, namely, the traditional vehicle platform, and three portable platforms, including handheld, wearable, and trolley-based systems. Further details of these systems are provided in Table 4 and the following subsections.

**Table 4.** Specifications of different MMSs.

| | System | Release Year | Indoor | Outdoor | Camera | LiDAR/Max. Range | IMU | GPS | Accuracy* | Applications |
|---|---|---|---|---|---|---|---|---|---|---|
| Vehicle-mounted | Leica Pegasus: Two Ultimate | 2018 | ✗ | ✓ | 360° FoV | ZF9012 profiler 360°x41.33°/100 m | ✓ | ✓ | 2 cm horizontal accuracy 1.5 cm vertical accuracy | • Urban 3D modeling. • Road asset management. • Analyzing change detection • Creating HD maps. • Generating geo-located panoramic images. |
| | Teledyne Optech Lynx HS600-D | 2017 | ✗ | ✓ | 360° FoV | 2 Optech sensors/130 m | ✓ | ✓ | ±5 cm absolute accuracy | |
| | Topcon IP-S3 HD1 | 2015 | ✗ | ✓ | 360° FoV | Velodyne HDL-32E Li-DAR/100 m | ✓ | ✓ | 0.1 cm road surface accuracy (1 sigma) | |
| | Hi-Target HiScan-C | 2017 | ✗ | ✓ | 360° FoV | 650 m | ✓ | ✓ | 5 cm at 40 m range | |
| | Trimble MX7 | | ✗ | ✓ | 360° FoV | | ✓ | ✓ | — | |
| | Trimble MX50 | 2021 | ✗ | ✓ | 90% of a full sphere | 2 MX50 Laser scanner/80 m | ✓ | ✓ | 0.2 cm (laser scanner) | |
| | Trimble MX9 | 2018 | ✗ | ✓ | 1 spherical + 2 side looking + 1 backward/downward camera | MX9 Laser scanner/up to 420 m | ✓ | ✓ | 0.5 cm (laser scanner) | |
| | Viametris vMS3D | 2016 | ✗ | ✓ | FLIR Ladybug5+ | Velodyne VLP-16 + Velodyne HDL-32E | ✓ | ✓ | 2-3 cm relative accuracy | |
| Handheld | HERON LITE Color | 2018 | ✓ | ✓ | 360° x 360° FoV | 1 Velodyne Puck/100 m | ✓ | ✗ | 3 cm relative accuracy | • Mapping enclosed and complex spaces and cultural heritage. • Forest surveying. • Building Information Modeling. |
| | GeoSLAM Zeb Go | 2020 | ✓ | ✗ | Can be added, accessory | Hokuyo UTM-30LX laser scanner/30m | ✗ | ✗ | 1 to 3 cm relative accuracy | |
| | GeoSLAM Zeb Revo RT | 2015 | ✓ | ✗ | Can be added, accessory | Hokuyo UTM-30LX laser scanner/30m | ✗ | ✗ | 0.6 cm relative accuracy | |
| | GeoSLAM Zeb Horizon | 2018 | ✓ | ✓ | Can be added, accessory | Velodyne Puck VLP-16/100m | ✗ | ✗ | 0.6 cm relative accuracy | |
| | Leica BLK2GO | 2018 | ✓ | ✓ | 3 camera system 300° x150° FoV | Up to 25 m 360x270 | ✗ | ✗ | ±1 cm in an indoor environment with a scan duration of 2 minutes | |
| Wearable | Leica Pegasus: Backpack | 2017 | ✓ | ✓ | 360° x 200° FoV | Dual Velodyne VLP-16/100 m | ✓ | ✓ | 2 to 3 cm relative accuracy<br>5 cm absolute accuracy | |
| | HERON MS Twin | 2020 | ✓ | ✓ | 360° x 360° FoV | Dual Velodyne Puck/100 m | ✓ | ✗ | 3 cm relative accuracy | |
| | NavVis VLX | 2021 | ✓ | ✓ | 360° FoV | Dual Velodyne Puck LITE/100 m | ✓ | ✗ | 0.6 cm absolute accuracy at 68% confidence<br>1.5 cm absolute accuracy at 95% confidence | |
| | Viametris BMS3D-HD | 2019 | ✓ | ✓ | FLIR Ladybug5+ | 16 beams LiDAR + 32 beams LiDAR | ✓ | ✓ | 2 cm relative accuracy | |
| Trolley | NavVis M6 | 2018 | ✓ | ✗ | 360° FoV | 6 Velodyne Puck LITE/100 m | ✓ | ✗ | 0.57 cm absolute accuracy at 68% confidence<br>1.38 cm absolute accuracy at 95% confidence | • Indoor mapping for government buildings, airports, and train stations.<br>• Tunnel inspection.<br>• Measuring asphalt roughness.<br>• Building Information Modeling. |
| | Leica ProScan | 2017 | ✓ | ✓ | ✗ | Leica ScanStation P40, P30 or P16 | ✓ | ✓ | 0.12 cm (range accuracy for Lecia ScanStation P40) | |
| | Trimble Indoor | 2015 | ✓ | ✗ | 360° FoV | Trimble TX-5, FARO Focus X-130, X-330, S-70-A, S-150-A, S-350-A | ✓ | ✗ | 1 cm relative accuracy when combined with FARO Focus X-130 | |
| | FARO Focus Swift | 2020 | ✓ | ✗ | HDR camera | FARO Focus Laser Scanner with a FARO ScanPlan 2D mapper | ✓ | ✗ | 0.2 cm relative accuracy at 10 m range<br>0.1 cm absolute accuracy | |

*The accuracy measurement reported by the manufacturers, the measure of the accuracy is unknown if not stated as relative or absolute. The "—" symbol indicates that the specifications were not mentioned in the product datasheet.

### 4.1. Vehicle-Mounted Systems

This setting refers to mounting the sensor suites on top of the vehicle to capture dense point clouds. These systems enable a high-rate data acquisition at the vehicle travel speed (20 to 70 miles/h). The sensor platform can be mounted on cars, trains, or boats depending on the mapping application. Generally, vehicle-mounted systems achieve the highest accuracy compared to other mobile mapping platforms primarily because of their size and payload, which allows them to host high-grade sensors [72]. A vehicle-mounted MMS system is usually equipped with a survey-grade LiDAR that provides dense and accurate measurements, as well as a deeply integrated 360° FoV camera providing textural information. Regarding the positioning sensors, a vehicle-mounted system usually fuses measurements from GNSS receivers with IMU and DMI. An example of these systems is introduced in [73] that uses one Velodyne HDL-32E, and five Velodyne VLP-16 LiDAR sensors combined with multiple GPS receivers and IMUs. Other examples include Leica Pegasus: Two Ultimate [13], Teledyne Optech Lynx HS600-D [74], Topcon IP-S3 HD1 [75], Hi-Target HiScan-C [76], Trimble MX50, MX9, MX7 [43], and Viametris vMS3D [77]. Figure 2 shows a sample of such vehicle-mounted systems.

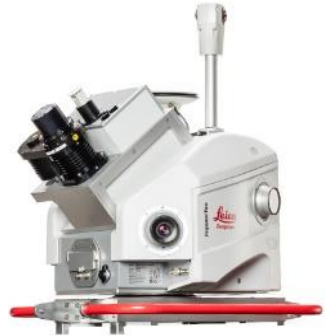

**Figure 2.** Leica Pegasus: Two Ultimate vehicle-mounted system. Photo courtesy of Leica Geosystems [13].

Vehicle-mounted systems are used for various applications such as urban 3D modeling, road asset management, and condition assessment [78,79]. Moreover, these systems can be used for automated change detection in the mapped regions [80,81], creating up-to-date HD maps as an asset for autonomous driving [73] and railway monitoring applications [82].

Although vehicle-mounted systems play a major role in mobile mapping, their relatively large size hinders their accessibility to many sites such as narrow alleys, and indoor environments. Additionally, some studies [83] have demonstrated that the speed of the vehicle may affect the quality of the 3D data creating doppler effects for successive scans [84]. Therefore, the speed and plan of the route have to be planned ahead of the mapping mission.

### 4.2. Handheld and Wearable Systems

The handheld and wearable systems follow lightweight and compact designs using small-sized sensors. An operator can hold or wear the platform and walk through the area of interest. Wearable systems are often designed as a backpack system to allow the operator to collect data while walking. Both handheld and wearable systems are distinguished by their portability, which enables mapping GPS-denied environments including enclosed spaces, complex terrains, or narrow spaces that vehicles cannot access [20,85]. Due to the nature of these environments, handheld and wearable systems may not rely on GNSS receivers for positioning, but instead, depend on an IMU or use a LiDAR and camera for both data collection and localization (using simultaneous localization and mapping (SLAM) approaches) [86]. A sample of handheld and wearable systems are shown in

Figure 3. Examples of these devices are HERON LITE Color [87], GeoSLAM Zeb Revo Go, Zeb RT, Zeb Horizon [88], Leica BLK2GO [13], Leica Pegasus: Backpack [13], HERON MS Twin [87], NavVis VLX [89], and Viametris BMS3D-HD [77]. Some other examples of the devices are introduced in [70,85,90], where they show the benefit of using LiDAR with IMU to generate 2D and 3D maps and evaluate the performance of these mapping devices in indoor environments.

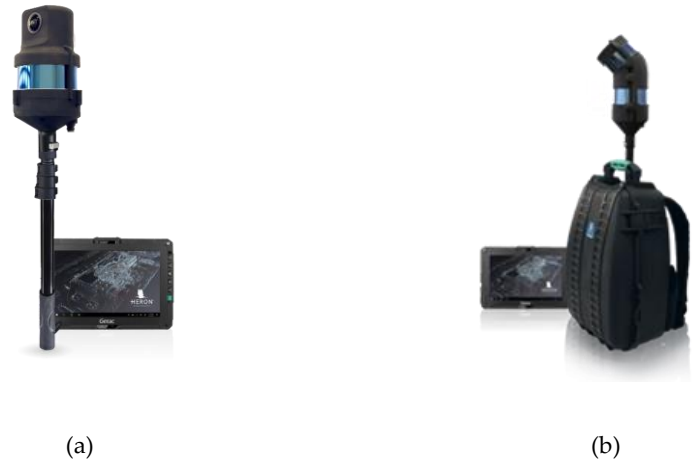

(a)                                         (b)

**Figure 3.** Handheld and wearable systems: (a) HERON LITE Color, (b) HERON MS Twin. Photos courtesy of Gexcel srl [87].

As mentioned above, handheld and wearable systems are effective in mapping enclosed spaces; for instance, these devices can be used to map caves where GNSS signals and lightings are not available [91]. In addition, they are used to map cultural heritage sites that may be complex and require the data to be efficiently collected from different viewing points [92-94]. Furthermore, these systems are efficient in mapping areas that are not machine-accessible, such as forest surveying [68,95,96], safety and security maps, and building information modeling (BIM) [97]. However, working in GPS-denied regions requires compensating for the lost signal, which demands setting GCPs in these regions, or utilizing GPS information before entering into such environments [98], whereas, inside the tunnels, the navigation completely depends on IMU/DMI or the scanning sensors (LiDAR or cameras).

### 4.3. Trolley-Based Systems

This type of system holds similar nature to the backpack systems while remaining to be slightly more sizeable to carry a heavier payload. It is suitable for indoor and outdoor mapping where the ground is flat [99]. A sample of trolley-based systems is shown in Figure 4. Examples of these systems include NavVis M6 [89], Leica ProScan [13], Trimble indoor [43], and FARO Focus Swift [100]. Trolley-based systems are also suitable for a variety of applications such as tunnel inspection, measuring asphalt roughness, creating floorplans, and BIM [22]. In addition, they are used for creating 3D indoor geospatial views of all kinds of infrastructure such as plant and factory facilities, residential and commercial buildings, airports, and train stations.

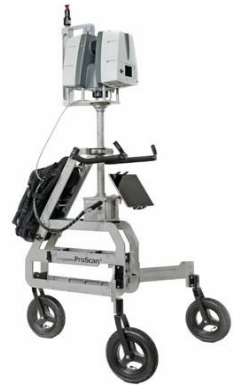

**Figure 4.** Leica ProScan trolley-based MMSs. Photo courtesy of Leica Geosystems [13].

## 5. MMS Workflow and Processing Pipeline

There are a few processing steps turning from the raw sensory data of MMS to the final 3D products, which generally include data acquisition, sensor calibration and fusion, georeferencing, and data processing in preparation for scene understanding (shown in Figure 5.). In the following subsections, we provide an overview of these typical processing steps.

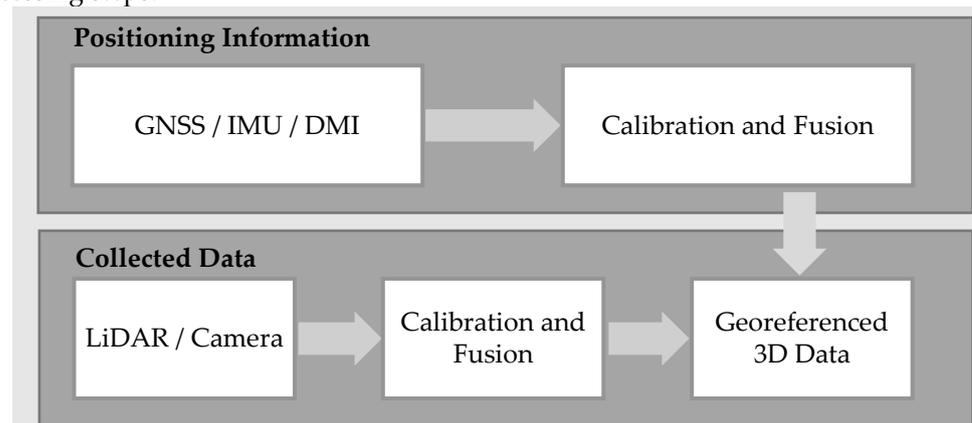

**Figure 5.** The standard processing pipeline for MMS data.

### 5.1. Data Acquisition

the planned route to be analyzed to determine the configured platforms and sensors to be deployed, for example, the operators should be aware of the GNSS accessible regions to plan the primary sensors to use. For MMS positioning, the GNSS, IMU, and DMI continuously measure the position and motion of the platform. In most outdoor applications, the main navigation and positioning data are provided by the GNSS satellite to the receiver, and the IMU and DMI supplements measurement where GNSS signals are insufficient or none. In some specific cases where GNSS is completely inaccessible, such as for cave mapping, GCPs will be used to reference the data to the geodetic coordinate system. 3D data are mainly collected by an integrated LiDAR and camera system, where LiDAR produces accurate 3D point clouds colorized by the images from the associated camera.

### 5.2. Sensors Calibration and Fusion

Sensor calibration and fusion are often performed throughout the data collection cycle, the task of which is to calibrate the relative positions between multiple sensors, including these between cameras, between camera and LiDAR, or among LiDAR, camera, and navigation sensors, as well as fusing their output as a postprocessing step to achieve more accurate positional measurements. These serve the purpose of both more accurate localization, more accurate geometric reconstruction, and data alignment for fusion [101,102]. In the following subsections, we introduce a few typical calibration procedures in MMS.

### 5.2.1. Positioning Sensors Calibration and Fusion

The integration of GNSS, IMU, and DMI is split into several steps. The first is lab/factory pre-calibration, which estimates the relative offset among these sensors and their relative position to the data collecting sensors (e.g., LiDAR and cameras) [103]. Second, sensor information fusion to output the estimated positions through optimal statistical/stochastic estimators [29,104]. A typically used algorithm is the Kalman filter (KF) [105-108], which uses continuous measurements over time with their uncertainties and a stochastic model for each sensor to estimate the unknown variables in a recursive scheme. KF is the simplest dynamic estimator that assumes linear models and Gaussian random noise of observations; thus is often re-adopted through Extended KF (EKF) for linearized non-linear models [109,110]. However, convergence is not guaranteed for EKF, especially when the random noise does not follow the Gaussian distribution. Thus, the particle filter [111,112] is usually adopted as a good alternative, as it can simulate the noises to deal with the potential non-Gaussian noise distributions.

### 5.2.2. Camera Calibration

Camera calibration refers to the process of rigorously determining the camera intrinsic parameters (i.e., focal length, principal point), various lens distortions (e.g., radial distortion), and other intrinsic distortions such as affinity or decentering errors [113,114]. The camera calibration parameters often follow the standard Brown model [115], or the extended parameters [116] that additionally model in-plane errors due to the film/chip displacements. The traditional and the most rigorous camera calibration approach uses a 3D control field consisting of highly accurate 3D physical point arrays. Converging images of these point arrays are captured at various angles and positions. These well-distributed 3D points, together with their corresponding 2D observations on the image, go through a rigorous BA with additional parameters (i.e., calibration parameters). However, 3D control fields are demanding and costly, thus this approach is mostly used for calibrating survey-grade or aerial cameras at the factory level. A popular and less demanding calibration approach is called cloud-based camera calibration [116] Instead of using the very expensive 3D control fields, this approach uses coded targets, which can be arbitrarily placed (but well-distributed with certain depth variations) in a scene as a "target cloud". Converging images of these targets can yield very accurate 2D multi-ray measurements, which are fed into a free-network BA for calibrating the camera parameters. Because of its simplicity, this is often used as a good alternative in calibrating cameras for close-range applications, including herein the MMS applications. A least rigorous but very often used calibration method uses a chessboard as a target for calibration [117], which extracts regularly distributed 2D measurements from images of the chessboard and performs self-calibrating BA. However, due to its limited scene coverage, the nature of the board being a planar thus lack of depth variation, as well as the limited flexibility to capture well-converged images filled with features, it may not, in BA, well decorrelate the camera parameters from the exterior orientation parameters leading to potential errors in calibration. Since it is very commonly used in the computer vision community and well supported by available open-source tools, it is one of the most popular approaches to obtain quick calibrations and can be used to calibrate cameras that do not demand high surveying accuracy, such as for navigation cameras.

When calibrating a multi-camera system (e.g., a stereo rig), the camera calibration is extended to additionally estimate the accurate relative orientation among these cameras. The calibration still undergoes through a BA, while it requires at least knowing the scale of the targets (either the "target clouds" or chessboard), in order to metrically estimate the baselines between all these cameras.

### 5.2.3. LiDAR and Camera Calibration

The calibration between LiDAR and the camera covers a few aspects: first, images must be time-synchronized with the LiDAR scans. Second, the relative orientation between the LiDAR and camera rays must be computed. Third, they must have the same viewpoint to avoid parallaxes. Time-synchronization is one of the crucial calibration steps to correct time offsets between sensors. It refers to matching the recorded and measured data from different sensors with separate clocks to create well-aligned data in terms of time and position. Time-synchronization errors (or the time offsets) are due to 1) clock offset, which refers to the time difference between internal clocks of sensors, or 2) clock drift, which refers to sensors' clocks operating at different frequencies or times [118]. A time offset greater than milliseconds between LiDAR and camera can cause significant positioning errors when recording an object. Additionally, the impact can be more significant and noticeable if the platform is operating at a high speed. To address this problem, sensors must have a common time reference, which is often based on GNSS's time because of its high precision and ability to record positions in nanoseconds [119]. The time offset between GNSS and IMU is either neglected because of its insignificance or maybe slightly corrected using KF as a fusion method. The LiDAR and camera timestamps are corrected in real-time using the computer system on board. The computer system updates that data based on GNSS's time, some examples of these computer systems/servers include GPS service daemon (GPSD), IEEE, and Chrony.

Relative orientation refers to estimating the translation and orientation parameters between sensors. This type of calibration often needs to be carried out once in a while, due to deteriorations of the mechanics within and among the sensors after operating in different environments. Since quick calibration can be performed using a single image, where four corresponding points are selected between the image and the 3D scan, using either the well-identified natural corner points or highly reflective coded targets.

Ideally, LiDAR and Camera data must be well-aligned, however because of the wide baseline between the two sensors they often view objects from different angles leading to large parallax between them. The large parallax causes crucial distortions such as flipping or flying points, flying points, occlusions, etc. To resolve this issue, the relative orientation parameters are used to project the LiDAR data into the camera coordinate system [120].

### 5.3. Georeferencing LiDAR Scans and Camera Images using Navigation Data

Both the LiDAR scan and images collect data in a local coordinate system, georeferencing them refers to determining their global/geodetic coordinates, mostly based on the fused GNSS/IMU/DMI positioning data. This is a process after the calibration among these data collection sensors (as described in Section 5.2). Georeferencing includes the estimation of the orientation (boresight) and position (lever-arm) parameters/offsets with respect to GNSS and IMU [19]. The boresight and level-arm parameters define the geometrical relationship between positioning and data collection sensors. There are two approaches to perform georeferencing: 1) the direct approach that uses just GNSS/IMU data, or 2) the indirect approach that uses GNSS/IMU data in addition to GCP and BA for refinement [68]. The direct approaches are less demanding since they do not require GCP, and they can achieve accuracy in the decimeter to centimeter levels. Indirect approaches can provide more accurate (centimeter-level) and precise results, whereas typical surveying methods like GCPs and BA are adopted, however, they are very expensive, and their accuracy may vary based on the GCP setup (i.e., position and number of GCPs).

### 5.4. Data Processing in Preparation for Scene Understanding

Mobile mapping is highly relevant to autonomous vehicles, where scene understanding is crucial to not only automate the mapping process but also provide critical scene information in real-time to support platform mobilizations. Scene understanding is the process of identifying the semantics and geometry of objects [121,122]. With the enhanced processing capability of mobile computing units, advanced machine learning models, and the ever-increasing datasets, there is a growing trend to perform on-board data processing and scene understanding using the collected measurements from the mobile system [123,124]. These include real-time detection, tracking, and semantic segmentation of both dynamic (e.g., pedestrians) and static (e.g., road markings or signs) objects in a scene [122,125]. This has driven the need to develop representative benchmark datasets, better-generalized training, and domain adaption approaches [126], and lighter machine learning models or network structures that support real-time result inferences [127], examples of these efforts include MobileNet [128], BlitzNet [127], MGNet [129] and MVLidarNet [130]. Challenges exist when addressing these needs, as the mobile platforms may collect data under extremely different illuminations (e.g., daylight and night), weather conditions (e.g., rainy, snowy, and sunny), as well as with drastically different sensor suites with different quality of raw data.

## 6. Applications

Mobile mapping provides valuable assets for different applications, driven by not only the broad availability of easy-to-use and portable MMS platforms but also their readiness under different operating environments. This is made particularly useful as most of these applications rely on regularly acquired data for detection and monitoring purposes, such as railway-based powerline detection/monitoring [131,132]. In this section, we review some of the main applications of mobile mapping technology including road asset management, conditions assessment, creating BIM, disaster response, and heritage conservations. Documented examples of these applications in publications are shown in Table 5 and are detailed in the following subsections.

**Table 5.** An overview of the selected mobile mapping applications.

| Selected Applications | | Highlights |
|---|---|---|
| Road asset management and condition assessment | Extraction of road assets [79]; road condition assessment [133]; detecting pavement distress using deep-learning [134]; evaluating the pavement surface distress for maintenance planning [135]. | • Vehicle-mounted system regularly operating on the road. <br> • More efficient than manual inspection. <br> • Leveraging deep learning to facilitate the inspection process. |
| BIM | Low-cost MMS for BIM of archeological reconstruction [136]; analysis of BIM for transportation infrastructure [137]. | • Data is collected with portable systems. <br> • Useful for maintenance and renovation planning. <br> • Rich database for better information management. |
| Emergency and disaster response | Network-based GIS for disaster response [138]; analyzing post-disaster damage [139]. | • Timely and accurate disaster response. <br> • Facilitates the decision-making process. <br> • Effective training and simulations. |
| Vegetation mapping and detection | Mapping and monitoring riverine vegetation [140]; tree detection and measurement [141-143]. | • Accurate and automatic measurements. <br> • Reduces occlusions for 3D urban models. |
| Digital Heritage Conservation | Mapping a complex heritage site using handheld MMS [92]; mapping a museum in a complex building [94]; numerical simulations for structural analysis of historical constructions [144]; digital heritage documentation [145]; mapping archaeological sites [146]; | • Utilizes the flexibility of portable platforms. <br> • Enables virtual tourism. <br> • Digital recording of cultural sites. |

development of a digital heritage inventory system [147].

**Road Asset Management and Condition Assessment:** MMS operating on roads can regularly collect accurate 3D data of the road and its surroundings, which facilitates road asset management, mapping, and monitoring (e.g., road signs, traffic signals, pavement dimensions) [79]. Creating road asset inventories is of great importance given the large volume of road assets. Furthermore, since the condition of the roads deteriorates over time, automatic means of regular transportation maintenance, such as pavement crack and distress detection are critically needed [133,135]. Therefore, a key benefit of generating an updated and accurately georeferenced road asset inventory is to allow automatic and efficient change detection, in place of traditionally laborious manual inspections [134].

Typically, road condition monitoring process consists of four steps [148]: 1) data collection using MMS, 2) defect detection, which can be performed automatically using deep learning-based approaches, 3) defect assessment, and 4) road condition index calculation to classify road segments based on the type and severity of the defect. Therefore, MMS data can further help in increasing the safety of the road, for example, by detecting road potholes [78,149], evaluating the location of speed signs before horizontal curves on roadways [150], or assessing the passing sight distance on highways [151].

**Building Information Modeling:** BIM is one of the most well-established technologies in the industry of architecture, engineering, and construction. It provides an integrated digital database about an asset (e.g., building, tunnel, bridge, or 3D model of a city) during the project's life cycle. Typical BIM stages include 1) rigorous data collection, preparation of the 2D plans, and uploading them to specialized software to convert them into a digital format. The collected data includes information such as architectural design (i.e., materials and dimensions), structural design (e.g., beams, columns, etc.), electrical and mechanical designs, sewage system, etc., 2) preparation of the 2D plans and 3) manual upload and update of the plans using specialized software. This can facilitate the design, maintenance, and renovation processes of engineering buildings and infrastructures. However, this can be a challenging task because of the large amount of data that needs to be collected and the lack of automated processes that can increase the time and cost.

Nowadays, MMSs have been widely adopted for BIM projects due to their high accuracy, time efficiency, and lower cost in collecting 3D data. The collected point clouds and images are used to produce the 3D reconstructed model of an asset, then processed under semantic segmentation or classification to extract detailed information of all elements in the asset, the final product is then transferred to the BIM software to extract and simulate important information related to the life cycle of the project. In general, MMS can provide sufficiently accurate results for the derived BIM products [20]. These derived products can be either 2D floor plans or 3D mesh or polyhedral models representing the structure of architectures, or the life cycle of the construction process [145]. A popular example of MMS in BIM is for 3D city modeling, where it can be used to collect information on roadside buildings [143-145] and their structural information like windows layout and doors [98,146]. Additionally, they can also be used to maintain plans and record indoor 3D assets and building layouts, which can be generated using handheld, backpack, or trolley MMSs.

**Emergency and Disaster Response:** The geospatial data provided by the MMSs are critical to improving emergency, disaster responses, and post-disaster recovery projects. MMSs provide cost and time-efficient solutions to collect and produce the 3D reconstructed models with detailed information about the semantics and geometry to navigate through an emergency or a disaster. MMSs have been reported to provide building-level information (e.g., floors, walls and doors) to a resolution/accuracy at the centimeter-level. In many cases, the building's plans are not up-to-date after construction, which may hinder the rescue mission in case of a fire emergency [139,152]. On the other hand, the MMS can provide an efficient alternative to produce an accurate and most updated 3D model of a facility or a building at a minimum cost, which facilitates emergency responses. Another example is collecting 3D data of the roadside assets and feeding them into GIS systems, which can serve as pre-event analysis tools to identify potential impacts of natural disasters through simulation (e.g., flood simulation, earthquake, etc.), in aid of making preventative plans ahead of time. The future directions of an MMS involve more efficient data collection, where it can be as simple as a man holding a cellphone imaging surroundings, and although this can be with low accuracy [3], it can supply critical geo-referenced information in a disastrous or emergency event to supply information for situation awareness and remedies.

**Vegetation Mapping and Detection:** Mobile mapping has shown great success in collecting high resolution and detailed vegetation plots, to create up-to-date digital tree inventories for urban green planning and management, which has greatly accelerated the traditionally laborious visual inspections [142]. In addition, vegetation monitoring is important to limit the decline in biodiversity and identify hazardous trees [153,154]. Therefore, these requirements impacted the advancement of keeping an up-to-date digital database for vegetation data. The collected 3D data can be used for modeling 3D trees for visualization purposes in urban 3D models. Typically, the workflow may consist of three main steps [155]: 1) tree detection by segmenting the generated point clouds, 2) simplifying the detected structure of point clouds, 3) deriving the geometry parameters such as canopy height and crown width and diameter [155,156]. Thus, the collected point cloud is used to detect trees, and low vegetation at the roadside [141,143] to take into account the occlusion on building facades to supply information for city modeling. Moreover, the collected data from MMSs can also be used in calculating urban biomass combatting the urban heat island effects to help in analyzing the influence of the ecosystem on climate change [157].

**Digital Heritage Conservation:** There is a growing trend that people realize the importance of digitally documenting archaeological sites in preserving cultural heritage [146,158,159], as many of these sites are in danger of deterioration and collapses, which may be accelerated due to extreme weather and natural disasters, such as the collapses of many cultural sites in Nepal or Iran due to earthquakes [160]. Therefore, there is a critical need to proactively document these sites when they are still in shape [92,161,162]. Moreover, a well-documented heritage site may enable other means of tourism such as visual tours, to off-load the site visit and reduce the human factors reducing the deterioration of these sites. As a means of collecting highly accurate 3D data, MMS has been used as one of the primary sources to create 3D models of complex and large heritage and archaeological sites. The data collection part for heritage sites often requires multi-scans for both the interior and exterior from different angles to generate occlusion-free and realistic 3D models. For example, a vehicle-mounted system can be used to drive around the sites to collect exterior information, and wearable/handheld devices can be used to scan the interiors of these sites [162].

*6.1. Summary*

In sum, we have discussed selected applications of mobile mapping that demonstrate the importance and necessity of utilizing MMSs in different scenarios. The adoption of mobile mapping technology in various applications has proven to not only increase productivity but also reduce the cost of operation. For instance, using digital assets for construction has led to a boost in productivity for the global construction sector by 14-15% [163]. In addition, digitizing historic structures through creating BIMs using mobile mapping data enables prevention conservation for heritage buildings which saves 40-70% of the maintenance cost [164]. Aside from the productivity and cost aspects, mobile mapping data pave the way for producing new road monitoring studies and methods that can increase road safety and dramatically reduce the probability of accidents [165].

**7. Conclusion**

### 7.1. Summary

In this paper, we have provided a thorough review of the state-of-the-art mobile mapping systems, their sensors, and relevant applications. We first reviewed sensors and sensor suites typically used in modern MMSs and discussed in detail their types, benefits, and limitations (Section 3). Then, we reviewed the mobile platforms including vehicle-mounted, handheld, wearables, etc., and we described in detail their collecting logistics, giving examples of modern systems of different types (Section 4), and specifically highlighted their supported use case scenarios. We further reviewed the critical processing steps that turn the raw data into the final mapping products (Section 5), including sensor calibration and fusion, and georeferencing. Finally, we summarize the most common applications (Section 6) using the capability of modern MMS.

### 7.2. Future Trends

Despite these many variations of MMSs and their sensor suites, the main goal of an MMS is to provide a means of collecting 3D data in close range, at possibly the maximal flexibility and minimal cost. Given the complex terrain environment, a single or a few MMS can hardly be sufficient at all levels of mobile mapping applications. Thus, although off-the-shelf solutions are partially available, developing or adapting MMS to designated applications is still an ongoing effort. So far, MMS is still regarded as an expensive collection means, as the involved equipment, sensors and manpower in handling the logistics and processing are still considerably great. Therefore, as far as we can conclude, the ongoing and future trends continue to be:

1) Reduced sensor cost for high-resolution sensors, primarily LiDAR systems with equivalent accuracy/resolution as currently used ones, but at a much low cost;
2) Crowdsourced and collaborative MMS using smartphone data, for example, the new iPhone version has already been equipped with a low-cost LiDAR sensor.
3) Incorporation of new sensors, such as ultra-wide-band tracking systems, as well as WiFi-based localization for use in MMS.
4) Enhanced (more robust) use of cameras as visual sensors for navigation.
5) Higher flexibility in sensor integration and customization as well as more mature software ecosystems (e.g., self-calibration algorithms among multiple sensors) to allow users to easily "plug-and-play" different sensors to match the demand for mapping different environments;
6) Advanced post-processing algorithms for pose estimation, data registration for a close-range scenario, dynamic object removal for data cleaning, and refinement for collections in the cluttered environment.
7) The integration of novel deep learning solutions at all levels of processing, from navigation and device calibration to 3D scene reconstruction and interpretation.

Given the complexity of MMSs and the application scenarios, a one-stop-shop solution arguably does not exist, however, it is possible to get the process of customization of the system easier if the above-mentioned challenges are consistently attacked. Our future work will encompass component-level surveys that provide the community with comprehensive views accelerating the convergence of solutions addressing the above-mentioned efforts.

**Author Contributions:** Conceptualization, M.E., H.A., and R.Q.; formal analysis, M.E. and H.A.; investigation, M.E. and H.A.; resources, M.E. and H.A.; writing—original draft preparation, M.E., H.A., and R.Q.; writing—review and editing, M.E., H.A., and R.Q.; visualization, M.E. and H.A.; supervision, R.Q. All authors have read and agreed to the published version of the manuscript

**Funding:** This research reflects efforts partially supported by ONR N00014-20-1-2141. Hessah Albanwan is sponsored by Kuwait University.

**Data Availability Statement:** Not applicable.

**Conflicts of Interest:** The authors declare no conflict of interest.

**Disclaimer:** Mention of brand names in this paper does not constitute an endorsement by the authors.